\documentclass[sigconf]{acmart} 
\setcopyright{none}
\renewcommand\footnotetextcopyrightpermission[1]{}
\acmConference[]{}{}{}
\acmBooktitle{}
\acmPrice{}
\acmISBN{}
\acmDOI{}

\usepackage{multirow}

\usepackage{amssymb}
\usepackage{graphicx}
\usepackage{tcolorbox}
\usepackage{ulem}
\newcommand{\xmark}{\textbf{\texttimes}}
\usepackage{xspace} 
\newcommand{\red}[1]{\textcolor{red!70!white}{#1}}
\newcommand{\green}[1]{\textcolor{green!80!black!70!white}{#1}}
\newcommand{\hide}[1]{}

\AtBeginDocument{%
  }

\begin{document}

\title{Interpretable Multimodal Zero-Shot ECG Diagnosis via Structured Clinical Knowledge Alignment}

\author{Jialu Tang$^{1}$, Hung Manh Pham$^{2}$, Ignace De Lathauwer$^{3}$, Henk S. Schipper$^{4}$, Yuan Lu$^{1}$, Dong Ma$^{2}$, Aaqib Saeed$^{1}$}

\affiliation{$^{1}$Eindhoven University of Technology \country{The Netherlands}}
\affiliation{$^{2}$Singapore Management University \country{ Singapore}}
\affiliation{$^{3}$Maxima Medical Center \country{The Netherlands}}
\affiliation{$^{4}$Erasmus Medical Center \country{The Netherlands}}

\begin{abstract}
Electrocardiogram (ECG) interpretation is essential for cardiovascular disease diagnosis, but current automated systems often struggle with transparency and generalization to unseen conditions. To address this, we introduce ZETA, a zero-shot multimodal framework designed for interpretable ECG diagnosis aligned with clinical workflows. ZETA uniquely compares ECG signals against structured positive and negative clinical observations, which are curated through an LLM-assisted, expert-validated process, thereby mimicking differential diagnosis. Our approach leverages a pre-trained multimodal model to align ECG and text embeddings without disease-specific fine-tuning. Empirical evaluations demonstrate ZETA's competitive zero-shot classification performance and, importantly, provide qualitative and quantitative evidence of enhanced interpretability, grounding predictions in specific, clinically relevant positive and negative diagnostic features. ZETA underscores the potential of aligning ECG analysis with structured clinical knowledge for building more transparent, generalizable, and trustworthy AI diagnostic systems. We will release the curated observation dataset and code to facilitate future research. Code released here: https://github.com/Tang-Jia-Lu/Zeta.
\end{abstract}

\maketitle

\section{Introduction}
Electrocardiogram (ECG) interpretation is a cornerstone of cardiovascular disease diagnosis, offering a fundamental, non-invasive assessment tool. Automated ECG analysis methods, particularly deep learning models, have achieved impressive performance on large-scale datasets~\citep{zhu2020automatic}, promising to enhance clinical workflows and support physician decision-making~\citep{johnson2025artificial}. However, a major barrier to the widespread adoption of these powerful end-to-end models in clinical practice is their lack of transparency and interpretability~\citep{gliner2025clinically}. Unlike the structured process of human ECG interpretation, which involves analyzing specific features like rhythms, conduction intervals, and morphological abnormalities, black-box AI models often provide predictions without a clear, clinically understandable rationale. This opacity undermines clinician trust and increases the cognitive load required to verify model outputs, a significant concern for both medical professionals and patients~\citep{rudin2019stop,lundberg2017unified}. While recent explainable AI (XAI) techniques for ECG have attempted to provide insights, such as highlighting salient signal regions or generating narrative reports~\citep{anand2022explainable,al2023explainable}, they often do not fully align with the structured differential diagnosis process that clinicians employ, where evidence for and against specific conditions is systematically weighed.

\begin{figure*}[!htbp]
\centering
  \includegraphics[width=0.95\textwidth]{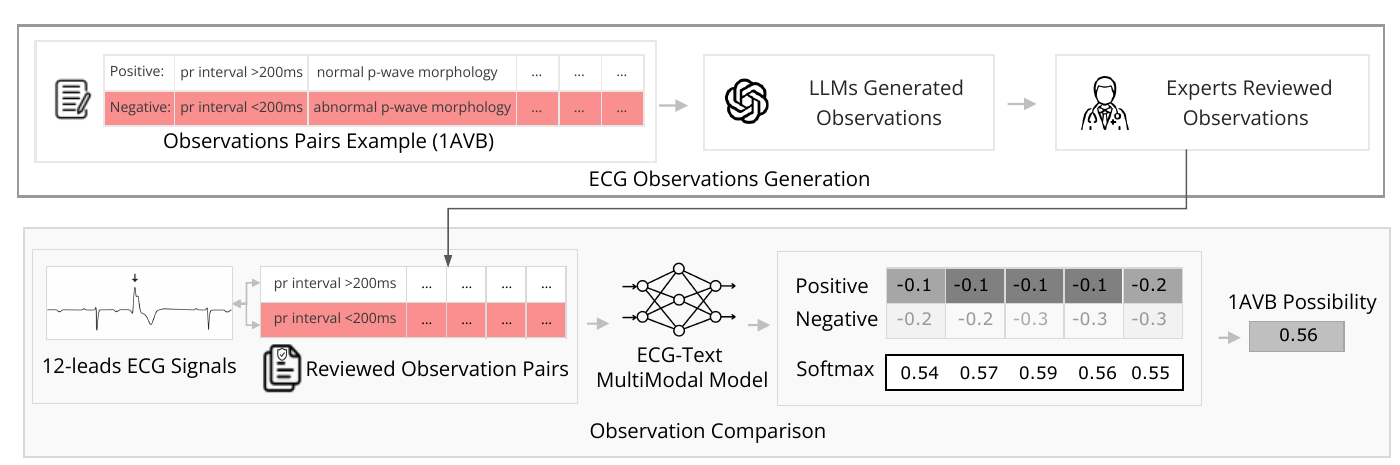}
  \caption{Overview of our ZETA framework. (1) ECG Observation Generation \& Curation: LLMs generate candidate positive/negative observations (e.g., for 1AVB), which are validated and refined by clinical experts. (2) Observation Comparison: A pre-trained multimodal model compares input ECG embeddings against expert-reviewed observation embeddings, yielding aggregated similarity scores for a final likelihood score. This process provides interpretable insights by linking predictions back to structured, clinically relevant diagnostic features.}
  \label{fig:teaser}
\end{figure*}

Furthermore, the clinical landscape is dynamic. Novel conditions may emerge, and the sheer volume and variability of ECG data make comprehensive expert annotation prohibitively expensive for training or fine-tuning of supervised models for every specific task or patient population. This motivates the exploration of approaches that can generalize to unseen conditions with minimal or no labeled data. Zero-Shot Learning (ZSL) is a promising paradigm that leverages semantic information from a self-supervised model to bridge the gap between seen and unseen classes, enabling recognition of novel categories without direct training examples~\citep{pourpanah2022review}. Existing ZSL methods applied to multimodal ECG analysis often rely on aligning ECG features with general textual descriptions of diseases~\citep{yu2023zero, pham2024c}. While promising for generalization, these approaches typically maintain an end-to-end structure where the link between specific ECG features and the final prediction is not explicitly interpretable in clinical terms. They lack the granular transparency needed to mirror the human diagnostic workflow and build clinical trust. Recent advances in multimodal learning and large language models (LLMs) offer new opportunities to bridge modalities and leverage rich textual knowledge, but effectively structuring this knowledge for interpretable, zero-shot medical diagnosis remains an open challenge.

In this paper, we propose ZETA (Zero-shot ECG Transparent Analysis), a zero-shot multimodal framework for ECG diagnosis that explicitly aligns with the structured clinical workflow of differential diagnosis, prioritizing interpretability alongside performance (see Figure~\ref{fig:teaser}). Our core idea is to leverage a pre-trained multimodal ECG-text model to compare ECG signals against \textit{structured clinical observations}. Crucially, these observations are organized as pairs of positive and negative diagnostic features associated with specific cardiovascular conditions. This comparison-based approach mirrors the clinical process of weighing evidence for (positive observations) and against (negative observations) a potential diagnosis, making the model's rationale intrinsically interpretable. To ensure the clinical relevance and quality of these observations, we propose a human-in-the-loop process involving LLM-assisted observation generation followed by validation and refinement by an expert. This collaborative approach creates a valuable dataset of structured ECG observations reviewed by experts, serving as the basis for our zero-shot comparison and interpretability evaluation.

We evaluate ZETA from multiple perspectives: its zero-shot classification performance compared to state-of-the-art methods, its generalization capabilities across different ECG datasets, and crucially, its interpretability and alignment with clinical utility. We demonstrate through qualitative examples and quantitative evaluation that ZETA provides a clinically grounded rationale for its predictions and that its performance aligns more closely with expert judgment compared to alternative approaches. Our main contributions are as follows:
\begin{enumerate}
    \item We propose ZETA, a novel zero-shot multimodal framework for interpretable ECG diagnosis that aligns ECG signals with structured positive and negative clinical observations, explicitly mimicking the differential diagnosis process.
    \item We develop a human-in-the-loop methodology involving LLM-assisted generation and expert review to create a curated dataset of expert-validated structured ECG observations. This process serves as a reference point for evaluating the interpretability and clinical relevance of AI-based ECG analysis.
    \item We provide extensive empirical evaluation demonstrating ZETA's competitive zero-shot performance and, show evidence of its improved interpretability and alignment with clinical intuition through qualitative analysis, addressing a key barrier to AI-based ECG analysis adoption.
\end{enumerate}
\section{Related Works}
\subsection{Explainable and Interpretable AI in ECG Diagnosis}
The development of interpretable AI models for ECG analysis is crucial for clinical trust and adoption. Existing XAI methods for ECG can broadly be categorized into several types. One approach involves generating synthetic ECG signals to illustrate decision boundaries or typical patterns associated with conditions, using techniques like GANs, VAEs, or diffusion models~\citep{berger2023generative,delaney2019synthesis}. While these methods can provide useful visualizations for understanding the model's learned representations, they do not directly provide a diagnosis or explain a specific prediction for a given patient's ECG in terms of clinical features.

Another common category focuses on identifying salient regions within the raw ECG signal that contribute most to a model's prediction. Techniques such as attention mechanisms~\citep{kuvaev2020attention}, saliency maps~\citep{jones2020improving}, and SHAP values~\citep{lundberg2017unified,bock2024enhancing} highlight specific time points or leads. This can help clinicians verify if the model is attending to relevant parts of the signal (e.g., a specific QRS complex or ST segment) based on their domain expertise. However, these methods only explain which parts of the signal are important, rather than why they are important in clinical terms (e.g., how a specific shape corresponds to a diagnostic finding like "ST elevation"). They typically rely on post-hoc analysis of a black-box predictor and do not incorporate structured medical knowledge or diagnostic rules directly into the prediction process.

Traditional ECG interpretation, in contrast, is inherently based on identifying and analyzing specific waveform morphology, intervals, and patterns according to established criteria and diagnostic rules~\citep{hancock2009aha}. Some prior methods have attempted to integrate rule-based reasoning or decision trees with traditional machine learning for interpretability~\citep{goettling2024xecgarch,kierner2023taxonomy}. While interpretable, these often require extensive manual rule engineering or may struggle with the complexity and variability inherent in real-world ECG signals compared to modern deep learning approaches. ZETA, unlike signal-saliency methods, integrates structured medical observations~\cite{pellegrini2023xplainer} (akin to diagnostic rules) directly into its core prediction mechanism and, unlike traditional rule-based systems, leverages powerful pre-trained multimodal representations.

\subsection{ECG-Language Multimodal Learning and Zero-Shot Learning}
Leveraging textual information alongside ECG signals has shown promise for enhancing diagnostic tasks and aligning AI model's outputs more closely with clinical documentation. ECG-Language models aim to bridge the gap between these two modalities. Some work has explored generating clinical reports from ECGs, which can provide human-readable summaries but often face challenges with accuracy, consistency, and generalization, especially when fine-tuned on limited data~\citep{fu2024cardiogpt}.

More relevant to our approach are methods that learn joint representations or alignments between ECG signals and medical text, often leveraging self-supervised or contrastive learning on large datasets. These models can support cross-modal retrieval and generation tasks. For instance, pre-trained models like D-BETA~\citep{pham2024c} learn strong embeddings by aligning ECGs with corresponding reports. While powerful for various tasks, the direct application of these models to diagnosis still lack transparency regarding \textit{why} a specific prediction was made in terms of clinically meaningful way.

Zero-Shot Learning (ZSL) has emerged as a technique to address the challenge of limited labeled data for rare or new conditions by leveraging semantic information. In the medical domain, ZSL can enable models to predict conditions they haven't been trained on by comparing patient data to textual descriptions of those conditions~\citep{pourpanah2022review}. Recent work has applied ZSL to ECG diagnosis by aligning ECG embeddings with embeddings of disease names or descriptive sentences derived from medical knowledge graphs or reports~\citep{yu2023zero}. 
While successful in achieving zero-shot generalization, these methods typically perform similarity comparisons at a high level (ECG embedding vs. disease description embedding). They often lack the granular interpretability needed for clinical validation–although the model outputs ECG finding (like "ST elevation in V2-V4" or "absence of P waves"), it does not perform clinician-like reasoning to clearly prioritize decisive findings that would allow verification of the prediction.

ZETA distinguishes itself by combining these threads. It leverages a pre-trained multimodal model for robust cross-modal understanding (like~\citep{pham2024c}), but applies this understanding in a novel ZSL framework that operates on \textit{structured positive and negative clinical observations}. This specific structure and comparison mechanism provides a direct link between the ECG signal and interpretable diagnostic features, explicitly mimicking the differential diagnosis process, a capability largely absent in existing ECG XAI and ZSL methods.
\begin{tcolorbox}[
    colframe=black,
    title=\footnotesize{LLM prompt for initial \textit{clinical observation} generation.}
]
\small{
Your job is to list observations cardiologists make when looking at ECG. Provided with a diagnosis, you have to extract key observations that are only relevant to the given diagnosis label. Delete observations that have the same symptoms and cannot be distinguished from a normal ECG. When generating observations, please do not include words like "suggest". For example, if an observation is "suggests ischaemia or left ventricular strain", you should generate: "ischaemia or left ventricular strain". 

\textbf{Diagnosis:} \texttt{\$\{Condition\}}  

Your output should be a JSON object with exactly 5 unique observations for both positive and negative cases:  
\begin{verbatim}
{
    "${Condition}": {
        "Positive": [...],
        "Negative": [...]
    }
}
\end{verbatim}
}
\label{box:prompt_template}
\end{tcolorbox}

\section{Method}
\label{sec:method}
We introduce ZETA (Zero-shot ECG Transparent Analysis), a novel framework designed for interpretable, zero-shot cardiovascular disease diagnosis from 12-lead ECG signals. ZETA operates by comparing ECG representations against structured, validated natural language descriptions of diagnostic observations. Our approach frames the zero-shot problem as a comparison between ECG signals and predefined sets of positive and negative textual observations associated with specific conditions. As illustrated in Figure \ref{fig:teaser}, the system comprises two main phases: (1) ECG Diagnostic Observations Generation and Curation, and (2) ECG-Language Aggregation Inference. The first phase focuses on creating high-quality, structured textual knowledge, while the second phase leverages a pre-trained multimodal model to perform inference by aligning ECGs with this structured knowledge.

\subsection{ECG Diagnostic Observations Generation and Curation}
\label{sec:observation_generation}
In this phase, we create a comprehensive set of structured clinical observations for various cardiovascular conditions. The core idea is to capture the key diagnostic features that cardiologists look for when interpreting an ECG, explicitly separating findings that support a diagnosis (positive observations) from those that argue against it (negative observations). This structure directly mirrors the clinical differential diagnosis process. To make this process scalable and leverage the power of state-of-the-art AI, we employ a expert-in-the-loop workflow involving Large Language Models (LLMs) and expert review.

\subsubsection{LLM-Assisted Observation Generation}
Leveraging the demonstrated capability of LLMs to understand and generate medical text~\cite{thirunavukarasu2023large}, we utilize them to pre-generate candidate diagnostic observations for target conditions. We designed a specific prompting strategy (see Box~\ref{box:prompt_template}) to guide LLMs to produce structured outputs. The prompt instructs the LLM to provide exactly distinct positive and five distinct negative observations for a given disease, formatted as a JSON object. Key instructions include focusing on disease-specific features, avoiding non-distinctive or vague findings, and ensuring clarity suitable for clinical interpretation. An example of LLM-generated observations for `Anterior Myocardial Infarction' condition is shown in Table~\ref{tab:ami_observations}. This step aims to generate a broad pool of potential observations efficiently, reducing the manual effort required from domain experts. 

\begin{table*}[htbp]
\centering
\caption{Example LLM-generated positive (P) and negative (N) ECG observations for Anterior Myocardial Infarction (AMI), prior to expert review.}
\begin{tabular}{@{}lll@{}}
\toprule
\textbf{Condition} &
\textbf{Positive Observations (P)} &
\textbf{Negative Observations (N)} \\ \midrule
AMI & loss of r waves v1-v4                              & normal r wave progression                                     \\
    & st-segment elevation in anterior leads             & normal st-segment in anterior leads                           \\
    & t-wave inversion in anterior leads                 & upright t waves in anterior leads                             \\
    & reciprocal st-segment depression in inferior leads & absence of reciprocal st-segment depression in inferior leads \\
    & reduced r wave amplitude in anterior leads         & normal r wave progression in inferior leads                   \\ \bottomrule
\end{tabular}
\label{tab:ami_observations}
\end{table*}

\begin{figure}[htbp]
\centering
\includegraphics[width=\columnwidth]{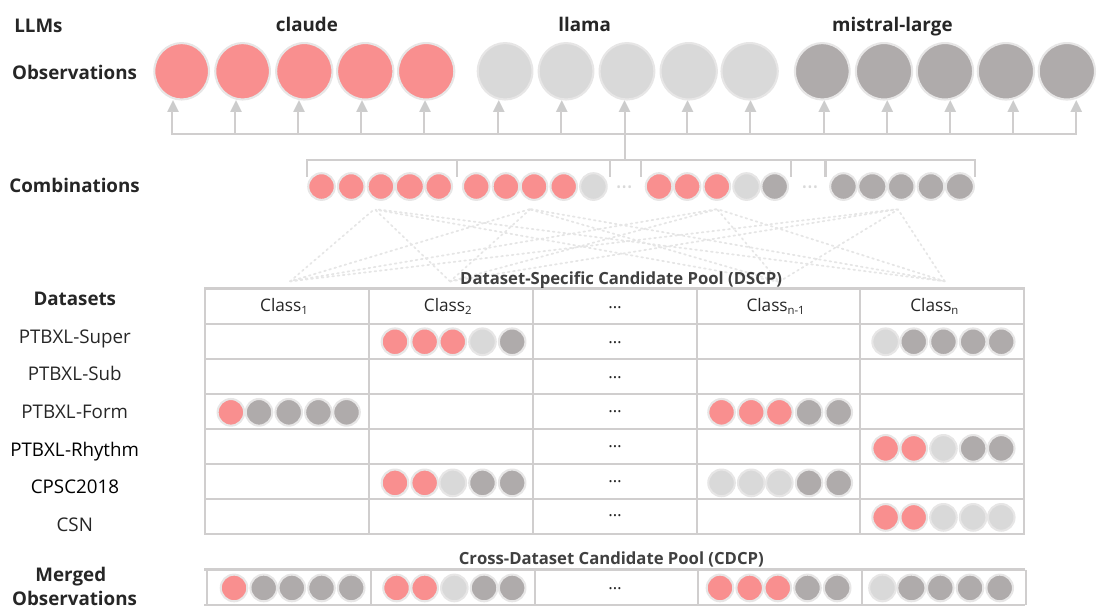}
\caption{Workflow for preprocessing LLM-generated clinical observations. Candidate observations from multiple LLMs are selected and organized into dataset-specific and cross-dataset sets for subsequent expert review.}
\label{fig:preprocess_generated_obs}
\end{figure}

\subsubsection{Preprocessing LLM-Generated Candidates}
\label{sec:preprocessing_observations}
After generating candidate observations from multiple LLMs (we selected Claude-3.5~\citep{anthropic2024a}, LLaMA-3.1~\citep{grattafiori2024llama}, and Mistral-Large-2~\citep{mistral2025} for this work based on their optimal balance of cost-effectiveness and performance capabilities at the commencement of our research), we perform preprocessing to prepare them for expert review and subsequent use in the multimodal model. For each condition, we collected 15 pairs of positive and negative observations (five from each of the three LLMs). To evaluate different strategies for utilizing these candidates, we created two sets:
\begin{itemize}
    \item \textbf{Dataset-Specific Candidate Pool (DSCP):} For evaluation on individual datasets (like PTB-XL-Form), we selected a specific set of candidate observations tailored to the conditions present in that dataset.
    \item \textbf{Cross-Dataset Candidate Pool (CDCP):} To evaluate generalization, we created a unified pool of candidates covering conditions present across multiple datasets (PTB-XL, CPSC2018, CSN). This involves merging and de-duplicating observations for overlapping conditions.
\end{itemize}
The goal of this preprocessing step is to create well-organized sets of candidate observations ready for expert validation, ensuring that the pool covers relevant features while managing redundancy. Figure~\ref{fig:preprocess_generated_obs} illustrates this process, showing how observations from different LLMs and datasets are combined and organized.

\begin{table*}[t]
\centering
\caption{Examples of expert review and modification of LLM-generated ECG observations for Sinus Rhythm (SR), Sinus Bradycardia (SBRAD), and Sinus Arrhythmia (SARRH), enhancing clarity, correctness, contrast, and directness. \red{Red} strike through text indicates removed or corrected content, while \green{green} highlights the revised observations.}
\begin{tabular}{@{}lllll@{}}
\toprule
\textbf{Condition} &          & \textbf{LLM Generated Observations}         & \textbf{Reasons} & \textbf{Reviewed}             \\ \hline
SR & Positive & regular pp intervals                   &            & regular pp intervals          \\
   &          & \red{\sout{consistent pr interval}}               & Directness & \green{normal hr 60-100}             \\
   &          & normal qrs \red{\sout{morphology}}                  & Clarity    & normal qrs \green{duration}       \\
   &          & normal p wave \red{\sout{morphology}}              &  Clarity    & normal p wave \green{before each qrs} \\
   &          & normal qt interval                     &            & normal qt interval             \\ \cline{2-5} 
   & Negative & \red{\sout{variable}} pp intervals                & Clarity    & \green{abnormal} pp intervals       \\
        &          & \red{\sout{variable pr interval}}            & Directness          & \green{hr \textless{}60 or \textgreater{}100} \\
   &          & \red{\sout{widened qrs complex}}                    & Clarity    & \green{abnormal qrs duration}        \\
   &          & \red{\sout{atrial fibrillation}}                  & Contrast   & \green{no distinct P waves}           \\
   &          & \red{\sout{ventricular tachycardia}}                 & Contrast   & \green{qt prolongation and shortening} \\ \hline
SBRAD   & Positive & \red{\sout{regular pp intervals}}           & Directness          & \green{hr \textless{}60bpm}               \\
   &          & \red{\sout{normal p wave morphology}}               & Directness   & \green{normal qrs}                   \\
   &          & \red{\sout{stable rhythm}}                         & Directness    & \green{consistent qrs morphology}     \\
   &          & normal p\red{\sout{-}}wave morphology               & Clarity    & normal p wave morphology       \\
   &          & \red{\sout{p-waves present and regular}}            & Directness &                                  \\ \cline{2-5} 
        & Negative & \red{\sout{pr interval fluctuation}}         & Directness          & \green{hr \textgreater{}60bpm}            \\
   &          & qrs widening                           &    & qrs widening                  \\
   &          & \red{\sout{st segment changes}}                     & Contrast    & \green{change in qrs morphology}      \\
   &          & \red{\sout{abnormal}} p\red{\sout{-}}wave morphology"              & Clarity    & \green{change in} p wave morphology  \\
   &          & \red{\sout{irregular r-r interval}}                 & Directness &                                  \\ \hline
SARRH   & Positive & changes correlate with respiration &              & Changes correlate with respiration    \\
   &          & irregular r\red{\sout{-}}r interval                    & Clarity    & irregular rr interval            \\
   &          &\red{\sout{ gradual acceleration/deceleration}}         & Directness &                                  \\
   &          & \red{\sout{heart rate within normal range}}           & Directness &                                  \\
   &          & \red{\sout{pr interval is normal (0.12-0.20 seconds)}} & Directness &                                  \\ \cline{2-5} 
   & Negative & \red{\sout{pr interval variation}}                     & Contrast    & \green{regular hrv}                      \\
   &          & regular r\red{\sout{-}}r interval                      & Clarity    & regular rr interval              \\
   &          & \red{\sout{irregular p wave morphology}}               & Clarity    &                                  \\
   &          & \red{\sout{bradycardia or tachycardia}}                & Directness &                                  \\
   &          & \red{\sout{normal qt interval}}                        & Directness &                                  \\ \hline
\end{tabular}
\label{tab:revision_observation}
\end{table*}

\subsubsection{Expert Review and Validation}
\label{sec:expert_review}
The generated observations undergo review and validation by a doctor (specializing in cardiology). This expert-in-the-loop step is critical for ensuring the clinical accuracy, relevance, and interpretability of the structured knowledge used by ZETA. The expert evaluates each candidate observation based on four key dimensions:
\begin{itemize}
    \item \textbf{Correctness}: Is the observation medically accurate and relevant to the specified condition?
    \item \textbf{Clarity}: Is the observation phrased clearly and unambiguously for clinical use?
    \item \textbf{Directness}: Does the observation take a decisive role in disease diagnosis or easily confused with other conditions?
    \item \textbf{Contrast}: For a given condition, do the positive observations clearly support the diagnosis, while the negative observations effectively argue against it or suggest an alternative?
\end{itemize}
Observations are revised or removed if they fail to meet these criteria. This expert-validated set of structured positive and negative observations forms the core knowledge base that ZETA uses for inference. Table~\ref{tab:revision_observation} provides examples of how LLM-generated observations were reviewed and modified by an expert to improve their quality based on these dimensions. The final, expert-reviewed observations are used exclusively in the subsequent inference phase.

\subsection{ECG-Language Aggregation Inference}
\label{sec:ecg_language_inference}
\begin{figure}[htbp]
\centering
\includegraphics[width=\columnwidth, angle=0]{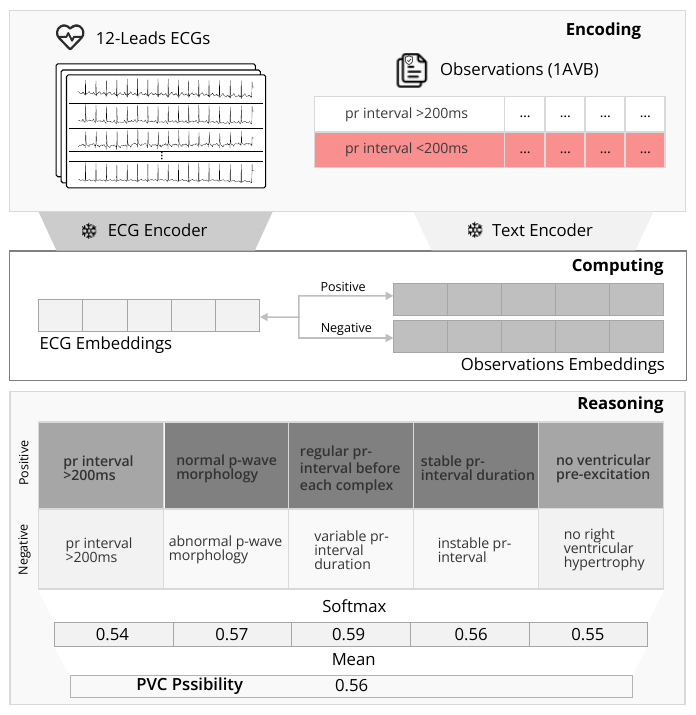}
\caption{ZETA's inference architecture: encoding ECG and P/N observations, computing similarities, and reasoning via observation scores to predict condition likelihood. Background color intensity represents ECG–observation similarity, with darker colors indicating higher similarity.}
\label{fig:model_structure}
\end{figure}

This phase describes how ZETA uses expert-validated structured observations to perform zero-shot diagnosis. The key idea is to calculate the similarity between the input ECG signal and the embedding of each positive and negative observation for a given condition.

\subsubsection{Multimodal Embedding Model}
\label{sec:pretrained_model}

To obtain robust, aligned representations of both ECG signals and text, we leverage a frozen pre-trained multimodal model, D-BETA~\citep{pham2024c}. D-BETA was trained on a large corpus of paired ECGs and clinical text reports using contrastive learning and masked modeling techniques. It consists of an ECG encoder (a transformer with convolutional layers) and a text encoder, linked by a cross-attention fusion module. By using this model \textit {in a frozen state}, we leverage its learned cross-modal embedding space without requiring task-specific fine-tuning on labeled data. This is fundamental to our zero-shot approach. The encoders produce 768-dimensional embeddings for both ECGs and text observations. We would like to highlight that our method is general-purpose in nature and can be used with any multimodal ECG-Language model. 

\subsubsection{Inference Process}
As depicted in Figure~\ref{fig:model_structure}, the inference process involves four steps: Encoding, Computing, Reasoning, and Classification.

\textbf{Encoding:} The input 12-lead ECG signal is passed through the frozen ECG encoder to obtain its embedding. Simultaneously, the text of each expert-reviewed positive and negative observation for the conditions of interest is passed through the frozen text encoder to obtain their respective embeddings.

\textbf{Computing:} For a given ECG and a specific condition, we compute the similarity between the normalized ECG embedding and the embedding of \textit{each} of its associated positive and negative observations. The similarity is calculated using the dot product between the ECG embedding vector and each observation's text embedding vector, both having 768 dimensions. A higher value of the dot product indicates a stronger alignment between the ECG features and the semantic content of the observation.

\begin{table*}[t]
\centering
\caption{\small{Zero-shot ECG classification performance (AUC) comparison across multiple datasets.}}
\resizebox{\textwidth}{!}{%
\begin{tabular}{lccccccc}

\hline
\textbf{Methods} & \textbf{PTBXL-Super} & \textbf{PTBXL-Sub} & \textbf{PTBXL-Form} & \textbf{PTBXL-Rhythm} & \textbf{CPSC2018} & \textbf{CSN} & \textbf{Average} \\ \hline
MERL~\citep{liu2024zero}   & 74.2 & 75.7 & 65.9 & 78.5 & 82.8 & 74.4 & 75.3 \\
D-BETA~\citep{pham2024c} & 76.2 & 75.9 & 66.1 & 88.6 & 80.1 & 76.3 & 77.1  \\ \hline
MERL~\citep{liu2024zero} (with ZETA)  &   70.0   &   61.9  &   59.8   &   60.0   &   69.0   &  62.8   &   63.9  \\ \hline
ZETA (DSCP) &  74.0 &   79.6   &   73.1   &   87.5   &   81.8   &   79.2   &   79.2   \\ 
ZETA (CDCP)  &   74.0   &   79.6   &   73.1   &   87.5   &   71.9   &   79.2   &   77.6  \\ 
ZETA (Expert-Reviewed)  &   74.0   &   81.0  &   73.9   &   90.2   &   78.1   &   72.4   &   77.2   \\ \hline
\end{tabular}%
}
\label{tab:zero-shot}
\end{table*}

\textbf{Reasoning:} This step aggregates the similarity scores to determine the overall likelihood of the condition being present, based on the evidence from the positive and negative observations. For a given condition, we have a set of similarity scores for its positive observations and a set for its negative observations. We apply a probability scaling (softmax with temperature $\tau=0.5$, following previous work~\cite{pellegrini2023xplainer}) to convert raw dot product scores into values between 0 and 1, representing the individual strength of the evidence for each observation. We then aggregate the scores for all positive observations (e.g., by taking the mean, as shown in Figure~\ref{fig:model_structure}) to get a single ``Positive Score'' for the condition, and similarly aggregate the scores for all negative observations to get a ``Negative Score''. The final ``Possibility Score'' for the condition is derived by comparing the aggregated Positive and Negative Scores (e.g., Positive Score - Negative Score, or using a logistic function on this difference). This comparison explicitly represents the weighing of evidence for and against the diagnosis, providing an interpretable link between the ECG features and the final diagnosis likelihood via the structured observations.

\textbf{Classification:} In the zero-shot setting, ZETA predicts the likelihood score for each potential condition by performing the Encoding, Computing, and Reasoning steps for the input ECG against the structured observations of each condition. For multi-label prediction, any condition whose ``Possibility Score'' exceeds a predefined threshold can be predicted. For evaluation purposes, the model's ability to discriminate between the true condition(s) and other classes is typically assessed using metrics like Area Under the Curve (AUC) on the Possibility Scores for each condition. The interpretability stems directly from the scores computed for each individual positive and negative observation during the Reasoning step), which can be presented to the clinician.

\section{Experimental Evaluation}
In this section, we present an empirical evaluation of ZETA to demonstrate its capabilities from several key perspectives: its zero-shot classification performance compared to state-of-the-art baselines, its generalization ability across different ECG datasets, and its alignment with clinical reasoning to provide interpretable diagnostic insights. Notably, our primary aim is not to surpass current classification metrics, but to demonstrate that ZETA achieves comparable performance to state-of-the-art methods while providing interpretability advantages by design.

\begin{table}[!hbp]
\centering
\resizebox{\columnwidth}{!}{
\small
\begin{tabular}{lcccccc}
\hline
\textbf{Model} & \textbf{1AVb} & \textbf{RBBB}   & \textbf{LBBB}   & \textbf{SB}     & \textbf{AF}  & \textbf{ST}  \\ \hline
Supervised DNN & 96.18       & 99.75 & 100.00 & 96.69          & 88.46 & 98.52           \\ 
Cardiology Resident & 83.80 & 98.21 & 95.00 & 96.69 & 88.28 & 90.48 \\ 
Emergency Resident  & 90.26 & 88.17 & 93.27 & 93.57 & 80.65 & 97.17  \\ 
Medical Student     & 95.36 & 96.87 & 94.87 & 87.25 & 95.60 & 91.70 \\ 
D-BETA~\cite{pham2024c}  & 94.33        & 95.31          & 96.55          & 97.26 & 98.17 & 99.11  \\ \hline
ZETA (Ours)                & 81.23 & 96.33 & 96.25 & 96.39 & 77.47 & 95.60 \\ \hline
\end{tabular}
}
\caption{Performance comparison of ZETA with Supervised DNN and Clinicians~\cite{ribeiro2020automatic}.}
\label{tab:code15}
\end{table}

Our evaluation utilizes several widely used public benchmark ECG datasets (with 12-leads) with multi-label annotations for various cardiac conditions, including PTB-XL \citep{wagner2020ptb}. We use different groupings (PTBXL-Super, PTBXL-Sub, PTBXL-Form, PTBXL-Rhythm) for evaluation as in prior work~\cite{pham2024c, liu2024zero}. CPSC2018 \citep{liu2018open}, CSN \citep{zheng2022large}, and CODE-15~\cite{ribeiro2020automatic} test set. These datasets cover a range of patient populations and 88 diagnostic conditions, providing a suitable basis for evaluating zero-shot performance and cross-dataset generalization. 

\begin{table*}[t]
\centering
\caption{Cross-dataset generalization performance under distribution shift. ZETA's interpretable zero-shot approach is compared against various SSL and other ZSL methods.}
\resizebox{\textwidth}{!}{%
\begin{tabular}{llcccccccc}
    \toprule
    \textbf{Source Domain} & & \multirow{2}{*}{\textbf{Zero-shot}} & \multirow{2}{*}{\textbf{Interpretable}} & \multicolumn{2}{c}{\textbf{PTBXL-Super}} & \multicolumn{2}{c}{\textbf{CSN}} & \multicolumn{2}{c}{\textbf{CPSC2018}} \\
    \cmidrule{0-0}\cmidrule(lr){5-6} \cmidrule(lr){7-8} \cmidrule(lr){9-10}
    \textbf{Target Domain} & & & & \textbf{CSN} & \textbf{CPSC2018} & \textbf{PTBXL-Super} & \textbf{CPSC2018} & \textbf{PTBXL-Super} & \textbf{CPSC2018CSN} \\
     
    \midrule
    SimCLR~\citep{chen2020simple} && \xmark & \xmark & 69.62 & 73.05 & 56.65 & 66.36 & 59.74 & 62.11 \\
    BYOL~\citep{grill2020bootstrap} & & \xmark & \xmark & 70.27 & 74.01 & 57.32 & 67.56 & 60.39 & 63.24 \\
    BarlowTwins~\citep{zbontar2021barlow} & & \xmark & \xmark & 68.98 & 72.85 & 55.97 & 65.89 & 58.76 & 61.35 \\
    MoCo-v3~\citep{chen2021empirical} &  & \xmark & \xmark & 69.41 & 73.29 & 56.54 & 66.12 & 59.82 & 62.07 \\
    SimSiam~\citep{chen2021exploring} &  & \xmark & \xmark & 70.06 & 73.92 & 57.21 & 67.48 & 60.23 & 63.09 \\
    TS-TCC~\citep{eldele2021time} &  & \xmark & \xmark & 71.32 & 75.16 & 58.47 & 68.34 & 61.55 & 64.48 \\
    CLOCS~\citep{kiyasseh2021clocs} &  & \xmark & \xmark & 68.79 & 72.64 & 55.86 & 65.73 & 58.69 & 61.27 \\
    ASTCL~\citep{wang2023adversarial} & & \xmark & \xmark & 69.23 & 73.18 & 56.61 & 66.27 & 59.74 & 62.12 \\
    CRT~\citep{zhang2023self} &  & \xmark & \xmark & 70.15 & 74.08 & 57.39 & 67.62 & 60.48 & 63.33 \\
    ST-MEM~\citep{na2024guiding} & & \xmark & \xmark & 76.12 & 84.50 & 62.27 & 75.19 & 73.05 & 64.66 \\
    MERL~\citep{liu2024zero} & & \checkmark &\xmark & 88.21 &78.01 & 76.77 & 76.56 & 74.15 & 82.86 \\
    
    D-BETA~\citep{pham2024c} & & \checkmark & \xmark & 72.09 & 79.11 & 77.12 & 82.91 & 76.24 & 80.10 \\
    
    \midrule
    ZETA (Ours) & & \checkmark & \checkmark & 79.10 & 80.64 & 75.13 & 83.98 & 73.99 & 72.30 \\
    \bottomrule
\end{tabular}
}
\label{tab:datashift}
\end{table*}

\subsection{Zero-Shot Classification Performance}
We evaluate ZETA's zero-shot performance, where the model has not seen any labeled examples of the target conditions during training. ZETA performs zero-shot prediction by calculating the similarity scores between an input ECG and the structured positive and negative observations for each potential condition, as described in Section~\ref{sec:ecg_language_inference}. Following prior work, we assess the performance using the Area Under the Receiver Operating Characteristic Curve (ROC AUC) per class.

Table \ref{tab:zero-shot} presents the zero-shot performance of different ZETA variants compared to existing zero-shot \hide{and few-shot} baselines, including MERL~\citep{liu2024zero} and D-BETA~\citep{pham2024c}, on various PTB-XL subsets, CPSC2018, and CSN. We evaluated three observation sets with ZETA. The evaluation encompasses 86 distinct conditions across 6 datasets, each represented by 2–5 observation pairs.
\textit{ZETA (DSCP and CDCP)}: Uses observations generated by individual LLMs (i.e., Dataset-Specific Candidate Pool) or a combined pool before expert review (Cross-Dataset Candidate Pool), demonstrating performance with potentially less refined knowledge.
\textit{ZETA (Expert-Reviewed)}: the expert-validated structured observations, representing the knowledge base aligned with clinical expertise.

The results show that ZETA (Expert-Reviewed) achieves competitive or superior zero-shot performance compared to other methods, particularly on the PTBXL-Rhythm subset where its performance is notably high (90.2 AUC). On PTBXL-Form, ZETA (Clinical-Reviewed) outperforms 
D-BETA~\citep{pham2024c}
by 7.8\%. While ZETA (Clinical-Reviewed) performs slightly lower on CPSC2018 compared to ZETA (DSCP), This may be attributed to the presence of a large number of repetitive sentences in the CPSC2018 dataset. While such redundancy may have been beneficial for model prediction, it negatively impacted clinicians’ ability to review and interpret patient conditions. As a result, these repetitive contents were removed, which in turn affected the model's final performance.
The overall average performance of ZETA (Clinical-Reviewed) (77.2 AUC) demonstrates its effectiveness as a zero-shot classifier leveraging structured clinical knowledge. Compared to MERL (with ZETA), using D-BETA improves performance by 13.3\%. This improvement is directly linked to D-BETA employs semantics-aware nearest-neighbor negatives to encourage the model to focus on distinguishing semantically similar yet unmatched ECG–text pairs, thereby enhancing the stability and robustness of contrastive learning.

To further contextualize ZETA's performance, we compare its ROC AUC scores against fully trained DNNs and human clinicians (Cardiology Resident, Emergency Resident, Medical Student) on a subset of common ECG classes where human performance data is available (1AVb, RBBB, LBBB, SB, AF, ST) from~\cite{ribeiro2020automatic}. Table \ref{tab:code15} shows this comparison. ZETA's performance (average 90.54) is competitive with Medical Students and Emergency Residents, though not reaching the level of a fully trained DNN or a Cardiology Resident. This comparison highlights that ZETA's zero-shot, interpretable approach provides a diagnostic capability within the range of human expertise levels, underscoring its potential utility in clinical workflows.

\subsection{Cross-Dataset Generalization}
We evaluate the generalization ability of ZETA and other self-supervised learning (SSL) methods under data distribution shifts. This is tested by training models on one source dataset and evaluating their performance on different target datasets. ZETA, leveraging a frozen pre-trained model (D-BETA) and text-based zero-shot inference, is inherently designed for generalization.

Table \ref{tab:datashift} presents the performance of various SSL methods and ZETA when trained on a source dataset (PTBXL-Super or CPSC2018) and tested on target datasets (CPSC2018, CSN, PTBXL-Super). The categories across these datasets, while not perfectly identical, have conceptual overlaps allowing for a meaningful cross-dataset evaluation. ZETA, using the Clinical-Reviewed observations and leveraging the generalization capability of the underlying D-BETA model, achieves on average similar performance across all tested distribution shifts. Notably, ZETA demonstrates strong transfer results, often outperforming several SSL methods, particularly when transferring from PTBXL-Super to CPSC2018 (80.64 AUC) and CSN to CPSC2018 (83.98 AUC). This indicates that aligning ECGs with structured clinical text provides a robust representation for generalizing across different data distributions while maintaining clinically grounded interpretability of the predictions.

\subsection{Evaluation of Interpretability and Clinical Alignment}
A core objective of ZETA is to provide interpretable predictions aligned with clinical reasoning. ZETA achieves this by providing prediction scores for each condition based on the aggregated evidence from positive and negative clinical observations. This allows users to see \textit{why} a particular condition is predicted or ruled out by examining the similarity scores for individual observations.

We evaluate ZETA's interpretability through qualitative analysis and its alignment with expert judgment during the observation curation process. As detailed in Section~\ref{sec:expert_review} and exemplified in Table \ref{tab:revision_observation}, the expert review process ensured the Correctness, Clarity, Directness, and Contrast of the clinical observations used by ZETA. This curation step is fundamental to guaranteeing that the model's ``reasons'' (the observations) are clinically valid and interpretable.

\begin{figure*}[h]
    \centering
    \includegraphics[width=\textwidth]{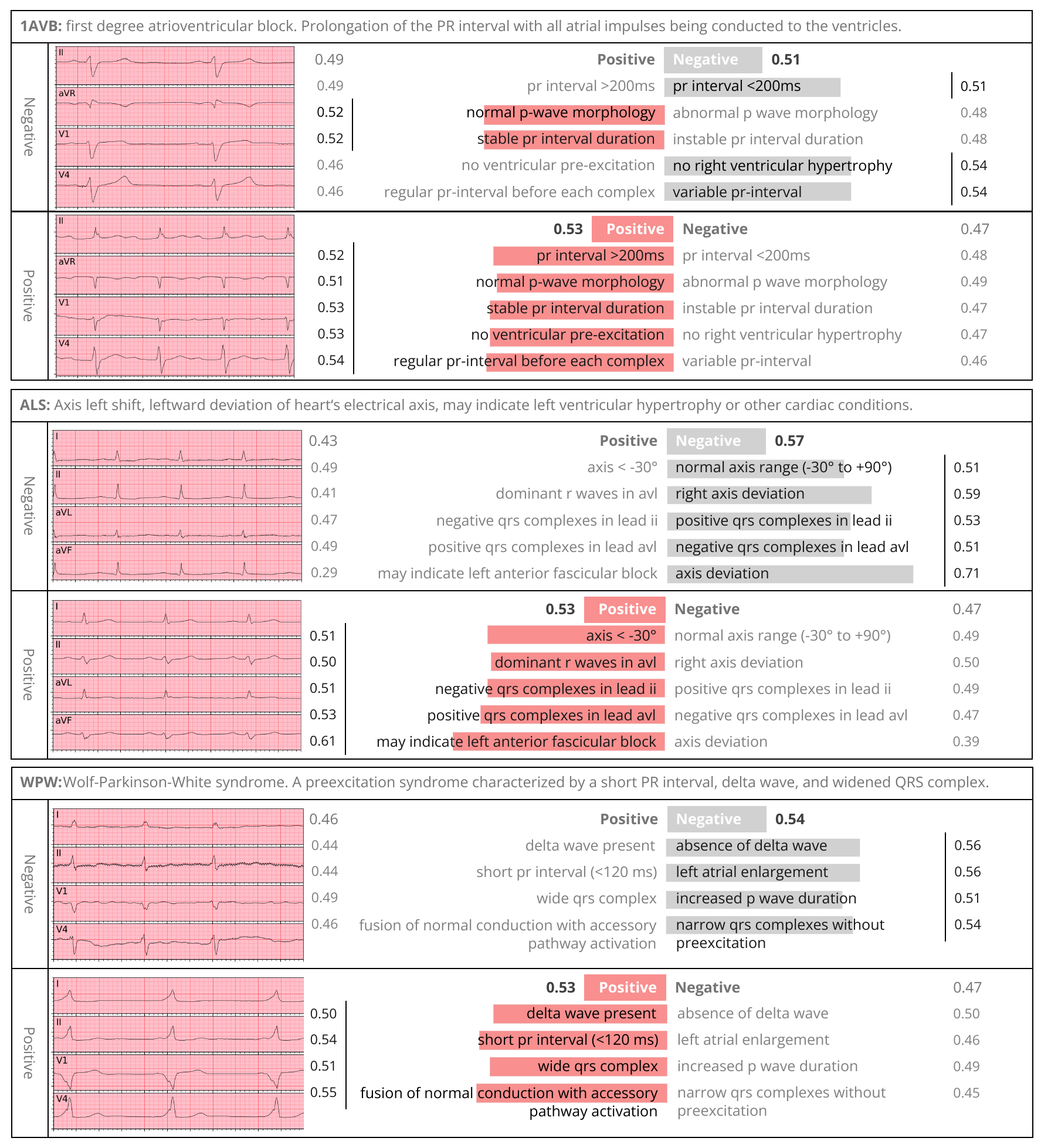}
    \caption{Qualitative analysis of ZETA's interpretable predictions for First Degree Atrioventricular Block (1AVB), Axis Left Shift (ALS), and Wolf-Parkinson-White syndrome (WPW). Scores indicate the alignment between the ECG (sample leads shown) and expert-reviewed P/N clinical observations, providing a transparent rationale for the final diagnosis likelihood.}
    \label{fig:qualitative_analysis}
\end{figure*}

Figure \ref{fig:qualitative_analysis} presents qualitative examples demonstrating ZETA's interpretability for three common conditions: 1AVB, ALS, and AFib. For each example ECG, we show the normalized similarity scores computed by ZETA for the corresponding positive and negative clinical observations (derived from the expert-reviewed set). For brevity we show only four ECG leads, the visualization shows which positive observations receive high scores (indicating features supporting the diagnosis are present in the ECG) and which negative observations receive low scores (indicating features arguing against the diagnosis are absent). For instance, in the 1AVB example, the ECG shows a high score for ``pr interval > 200ms'' (Positive Observation) and a low score for ``pr interval < 200ms'' (Negative Observation), directly linking the model's prediction to a key diagnostic criterion visible in the ECG. This level of detail provides clinicians with a transparent view into the model's reasoning process, allowing them to verify the prediction based on observable ECG features and validated clinical knowledge, thereby building trust. This evaluation effectively illustrates how ZETA translates its internal multimodal reasoning into clinically meaningful and verifiable evidence. 

The process of generating and reviewing observations (Section~\ref{sec:observation_generation}), also highlighted challenges and refinements necessary to align LLM outputs with clinical needs. The review process revealed issues like redundancy or insufficient contrast in initial LLM outputs, necessitating expert modifications to enhance the discriminative power and clarity of the observations used by ZETA. This underscores the value of the human-in-the-loop curation for building a robust and interpretable system. While Table \ref{tab:code15} primarily shows performance, the fact that ZETA performs comparably to human experts using this interpretable approach suggests the structured knowledge and comparison mechanism are effective in capturing clinically relevant information.

\subsection{Evaluating Interpretability in Clinical Decision Support}

To rigorously assess the interpretability of ZETA and its utility in a clinical decision support context, we conducted a human-centered study. The primary objective was to determine whether the structured positive and negative observations provided by ZETA can (1) effectively aid clinicians in their diagnostic process, (2) reflect meaningful medical reasoning, and (3) identify potential risks of clinicians being misled by the model's explanations, particularly when ZETA's underlying prediction is incorrect.

The study utilized the samples from PTB-XL dataset, focusing on four common rhythm-related diagnostic classes: Sinus Tachycardia (STACH), Atrial Flutter (AFLT), Atrial Fibrillation (AFIB), and Paced Rhythm (PACE). For each of these four classes, we selected 12 ECG samples: 6 with a positive ground truth label for the target class and 6 with a negative ground truth label. Within each set of 6, we chose 3 samples for which ZETA (Clinical-Reviewed) had the highest prediction scores for the target class (representing high-confidence predictions) and 3 samples for which ZETA had the lowest prediction scores (representing low-confidence predictions). This stratified sampling strategy aimed to present a balanced set of cases, including instances where ZETA's reasoning was likely helpful and instances where it might be potentially misleading.

Two physicians (referred to as `experts` hereafter) participated in the evaluation. In a blinded setting, for each ECG sample, experts were presented with the raw 12-lead ECG waveform alongside ZETA's corresponding structured positive and negative observations for the target rhythm class. Crucially, experts were \textit{not} shown final prediction score or the ground truth label for the ECG. Based solely on the raw ECG and ZETA's structured observations as diagnostic hints, each expert independently provided their diagnosis (presence or absence) for the target rhythm class for each ECG.

To evaluate the interpretability of our method, whether its explanations are understandable and actionable, we measured expert's diagnostic accuracy. This allows us to assess whether the model's highlighted observations intuitively support them in making correct decisions. Our designed study assesses both the supportive and robust qualities of the model’s observations. Specifically, we measured physician accuracy under two distinct conditions: when the model's prediction was correct (guidance) and when it was incorrect (misleading). This setup allows us to examine whether the model's highlighted observations genuinely help users make correct decisions when appropriate, and equally importantly, whether they avoid misleading them when the model is wrong. By testing interpretability in both favorable and unfavorable scenarios in Table \ref{tab:physician_performance}, we assess not only how useful the explanations are but also how robust and trustworthy they remain in error-prone settings.

\begin{table}[ht]
    \centering
    \caption{Physician performance with ZETA's structured observations.}
    \begin{tabular}{lcccc}
        \toprule
        Condition & Accuracy & Sensitivity & Specificity & F1-score \\
        \midrule
        Guidance   & 86.2 & 86.2 & 61.6 & 85.7\\
        Misleading & 87.3 & 87.3 & 47.8 & 87.0 \\
        \bottomrule
    \end{tabular}
    \label{tab:physician_performance}
\end{table}

To further evaluate interpretability in practical setting, we compared model-alone performance with physician decisions made under the guidance of model's observations. This comparison allows us to assess collaborative interpretability—whether the explanations can enhance human decision-making beyond what the model achieves on its own. An improvement in physician accuracy under this setting indicates in Table \ref{tab:model_vs_physician} that the observations are not only understandable, but also actionable, supporting their use in real-world clinical workflows where human oversight remains essential.

\begin{table}[ht]
    \centering
    \caption{Comparison of model-only and physician-assisted performance.}
    \resizebox{\columnwidth}{!}{
    \begin{tabular}{lcccc}
        \toprule
        Setting & Accuracy & Sensitivity & Specificity & F1-score \\
        \midrule
        ZETA                & 71.6 & 71.6 & 78.3 & 75.0 \\
        Expert + ZETA   & 86.5 & 86.5 & 58.7 & 86.0 \\
        \bottomrule
    \end{tabular}
    }
    \label{tab:model_vs_physician}
\end{table}

\section{Discussion}
This work introduced ZETA, a framework that uniquely bridges the gap between zero-shot learning (ZSL) and clinically-aligned interpretability for ECG diagnosis. Our results demonstrate that by leveraging structured positive and negative clinical observations, ZETA achieves competitive zero-shot performance and robust cross-dataset generalization without requiring task-specific fine-tuning. The core contribution lies in its transparent reasoning process, which mimics the differential diagnosis workflow by explicitly weighing evidence for and against a condition. This provides a clear, interpretable rationale for each prediction, directly addressing the ``black-box'' problem that hinders the clinical adoption of many high-performing deep learning models.

A key implication of our work is the human-in-the-loop methodology for knowledge curation. While the pursuit of higher accuracy often dominates machine learning research, building trust in clinical AI requires a focus on interpretability, which necessitates expert involvement. Our approach uses LLMs as a ``knowledge accelerator'' to generate candidate observations, significantly reducing the manual burden on clinicians who can then focus on the high-value task of validation and refinement. This collaborative model demonstrates a practical pathway for creating the structured, high-quality knowledge bases required for building next-generation interpretable AI systems, not only for ECG analysis but potentially for other medical modalities.

Despite its promise, ZETA has several limitations that define clear avenues for future research. First, its performance is fundamentally tied to the capabilities of the underlying pre-trained multimodal model. Any biases or weaknesses in the model's ability to represent ECG or text features are inherited by our framework. Second, the quality and comprehensiveness of the expert-reviewed observation set are paramount; the system is only as good as the clinical knowledge it is given. While our methodology streamlines curation, it remains a careful, expert-driven process.

Finally, ZETA's current interpretability is text-based. It explains what diagnostic feature the model identified as relevant, but does not yet show where in the signal that feature is located. To address this, the most immediate future work could focus on integrating signal-level saliency techniques. By generating attribution maps that highlight the specific regions of the ECG corresponding to the highest-scoring positive observations, it can create a powerful dual-explanation system. This would allow clinicians to not only read the model's rationale but also visually verify its findings on the ECG trace, further enhancing trust and clinical utility. The exploration of integrating more advanced multimodal foundation models as they become available to further improve performance and robustness is also an avenue for further exploration.

\bibliographystyle{ACM-Reference-Format}
\bibliography{main_ref}
\end{document}